\documentclass{article}

\usepackage[final]{neurips_2022}

\usepackage[utf8]{inputenc} 
\usepackage[T1]{fontenc}    
\usepackage{hyperref}       
\usepackage{url}            
\usepackage{booktabs}       
\usepackage{amsfonts}       
\usepackage{nicefrac}       
\usepackage{microtype}      

\usepackage[dvipsnames]{xcolor}
\usepackage{graphicx}
\usepackage{amsmath}
\usepackage{amssymb}
\usepackage{booktabs}
\usepackage{textcomp}
\usepackage{multirow}
\usepackage{array}
\newcolumntype{P}[1]{>{\centering\arraybackslash}p{#1}}
\newcolumntype{M}[1]{>{\centering\arraybackslash}m{#1}}
\usepackage{caption}
\usepackage{subcaption}

\usepackage{bm}
\def \bmx{{\bm{x}}}
\def \bmy{{\bm{y}}}

\def \bmphi{{\bm{\phi}}}

\definecolor{blue}{HTML}{4C72B0}
\definecolor{orange}{HTML}{DD8452}
\definecolor{green}{HTML}{55A868}
\definecolor{purple}{HTML}{8172B3}
\definecolor{brown}{HTML}{937860}

\usepackage{arydshln}

\title{PatchRot: A Self-Supervised Technique for Training Vision Transformers}

\author{%
  Sachin Chhabra\\
  Arizona State University\\
  Tempe, AZ, USA \\
  \texttt{schhabr6@asu.edu} \\
  \And
   Prabal Bijoy Dutta\\
  Arizona State University\\
  Tempe, AZ, USA\ \\
  \texttt{pdutta6@asu.edu} \\
  \And
   Hemanth Venkateswara\\
  Georgia State University\\
  Atlanta, GA, USA \\
  \texttt{hvenkateswara@gsu.edu} \\
   \And
   Baoxin Li\\
  Arizona State University\\
  Tempe, AZ, USA\\
  \texttt{baoxin.li@asu.edu} \\
  \AND
}

\begin{document}

\maketitle

\begin{abstract}
Vision transformers require a huge amount of labeled data to outperform convolutional neural networks. However, labeling a huge dataset is a very expensive process. Self-supervised learning techniques alleviate this problem by learning features similar to supervised learning in an unsupervised way. In this paper, we propose a self-supervised technique PatchRot that is crafted for vision transformers. PatchRot rotates images and image patches and trains the network to predict the rotation angles. The network learns to extract both global and local features from an image.  Our extensive experiments on different datasets showcase PatchRot training learns rich features which outperform supervised learning and compared baseline.
\end{abstract}

\section{Introduction}
\label{sec:intro}
In the past decade, convolution neural networks (ConvNets) have made tremendous progress in various image processing fields like object recognition (\cite{szegedy2015going,he2016deep}), segmentation (\cite{ronneberger2015u}), etc. This progress is primarily a result of supervised training on labeled data. The features learned by such a network are highly transferable and can be used for similar tasks (\cite{pan2009survey,chhabra2021glocal,wang2018deep}). However, labeling a dataset is generally expensive and time-consuming. 

Self-supervised learning alleviates this problem by learning rich features without the need for manual annotation of labeling the dataset. Hence, training a network in an unsupervised way is gaining a lot of traction. 
Self-supervised learning techniques train a network by creating an auxiliary task that requires an understanding of the object. 
(\cite{noroozi2016unsupervised}) divided an image into patches and trained the network to solve a jigsaw puzzle of the shuffled patches. Another technique predicts the angle of a rotated image (\cite{gidaris2018unsupervised}).

Existing self-supervised techniques were designed to train ConvNets. However, in recent years, Vision Transformers (ViT) have surpassed ConvNets (\cite{dosovitskiy2020image}). Transformers were initially designed for Natural Language Processing (\cite{vaswani2017attention}) but now are being applied to other modalities like speech (\cite{li2019neural}), image (\cite{dosovitskiy2020image}), video (\cite{arnab2021vivit}) etc. Nevertheless, ViTs are considered data-hungry models as they outperform ConvNets only when huge labeled training data is available. When the amount of training data available is limited, their performance is not as good as ConvNets. This is primarily because they lack the inductive bias of ConvNets like translation equivariance and locality (\cite{dosovitskiy2020image}). This elevates the importance of unsupervised training of ViTs. 
Existing self-supervised techniques can be applied to ViTs. 
However, we need to keep in mind that ViTs process images differently than ConvNets. ConvNets process the image like a grid and learn shared kernels to extract features whereas ViTs divide the images into patches and apply self-attention to the embedding of the patches. 

In this paper, we propose PatchRot, a self-supervised technique for ViTs. PatchRot trains a ViT to predict the rotation of patches as well as that of the image. Given an image, we either rotate the image or its patches at a time. The classification head of the ViTs contains the global information of the image and the patch heads contain the local information. The classification head is used to predict the rotation angle of the image and the other heads are used to predict the rotation angle for their respective patches. This way, PatchRot learns to extract both global and local features.
PatchRot is a simple self-supervised technique crafted for ViTs. It trains the classification head as well as the patch heads. Our experiments on different datasets showcase the strength of PatchRot over supervised training and compared baselines.

\section{Related Work}
\label{sec:relatedwork}
\subsection{Self-Supervised Learning}
Self-supervised learning methods aim to learn rich features in an unsupervised way that can be directly used in other tasks (downstream tasks) such as image classification, object detection, image segmentation, etc. 
These techniques generally design a surrogate task that can generate training supervision without manual annotation and solving them would require an understanding of the object.
(\cite{Zhang2016ColorfulIC}) proposed to train a ConvNet to colorize a grayscale image. 
(\cite{pathak2016context}) trained a network to fill the missing parts of an image.
(\cite{Doersch2015UnsupervisedVR}) trained the network to predict the relative position of a patch given another patch from the same image.
(\cite{noroozi2016unsupervised}) extracted nine patches from an image and trained the network to solve the jigsaw puzzle of the shuffled patches.
(\cite{gidaris2018unsupervised}) proposed a pretext task of predicting the angle of 2D rotation ($0$\textdegree$, 90$\textdegree$, 180$\textdegree$, 270$\textdegree) applied to an input image to generate features that are supervised through this task. 
\cite{Dosovitskiy2016DiscriminativeUF} proposed a training scheme to learn features by trying to predict the classes of samples generated by a set of transformations. 

\subsection{Vision Transformers}
In 2020, (\cite{dosovitskiy2020image}) introduced a new class of image-processing networks - Vision Transformers (ViT) that operates solely on the attention mechanism. Next, we discuss the different components of a ViT - Image Tokenization, Patch Embedding block, Encoder block, and MLP head.

\textbf{Image Tokenization}: 
ViT splits the 2-dimensional image using into $16 \times 16$ non-overlapping image patches in top-left to bottom right order. This forms a 1-dimensional sequential input of image patches which are then used as the input.

\textbf{Patch Embeddings}: 
A fully connected layer is used to convert each image token into an embedding vector. Splitting the image into 1-D image patches destroys the spatial relationships between the patches. Hence, a learnable spatial embedding is added to the embedding of patches to provide spatial information. Similar to BERT (\cite{devlin2018bert}), a learnable class token is added to the start of the patch embedding sequence and is used for image classification later.

\textbf{Transformer Encoder}:
An encoder block consists of a multi-headed self-attention block and a forward expansion block. Self-attention block computes attention between the input tokens and the forward expansion block utilizes multiple fully-connected layers to transform the input. Both the blocks use a normalization layer before processing the input. A residual connections are used between input, self-attention block output and the  forward expansion block.

\textbf{MLP Head}:
At the last encoder block, the output of the class token (added in patch embedding block) is used and the rest of the outputs are ignored. The class token output is processed through a multi-layer perception classifier (also known as MLP head) which classifies the image into one of the categories.

\section{Methodology}
\subsection{Problem Definition}
Let $\bmx \in [0,1]^{C\times H\times W}$ be the input image, where $C$, $H$, $W$ represent the number of channels, the height, and the width of the image, respectively. The goal is to train a ViT $T_\bmphi$ with parameters $\bmphi$ in an unsupervised manner and extract high-quality features. 
As input to the ViT, we preprocess the image into a sequence of image patches $s(\bmx)= [\bmx^s_1, \bmx^s_2, ..., \bmx^s_N]$, where $N$ is the number of patches. 
The image patch dimensions are determined by patch size $P$ where $\bmx^s_i \in [0,1]^{C\times P\times P}$ represents the $i$-th patch of image $\bmx$ and $N$ = $\frac{H}{P}\cdot\frac{W}{P}$. 
The image height $H$ and width $W$ is a multiple of patch size $P$ to ensure that $N$ is an integer.

\subsection{PatchRot}
\label{method}
PatchRot is a technique to train a ViT where the rotation angle of an image and the rotation angle of individual patches of an image are predicted.
RotNet (\cite{gidaris2018unsupervised}) showcased that a ConvNet trained to predict the rotation of an image learns features as good as those learned using supervised training. 
We train a ViT where the classification head (generally used for predicting the object category) is used to predict the rotation angle of the image.
We use the last encoder output of the other heads to predict the rotation angles for the individual patches using new multilayer perceptron (MLP) heads.
This way the ViT can produce an output for every element in the input sequence whereas ConvNets are limited to just one output. Therefore, PatchRot applies only to ViTs.

We represent the rotation operation as $R(;\iota)$, where $\iota \in \{0, 1, 2, 3\}$ that rotates the input by $\theta = 90$\textdegree.$\iota$ resulting in rotation of $0$\textdegree, $ 90$\textdegree$,$ $180$\textdegree, or $270$\textdegree.
A rotated image $\bmx_r = R(\bmx; \iota_0)$ uses $y=\iota$ as its the label for training.
Similar to (\cite{gidaris2018unsupervised}), we found that training using all 4 angles of rotations in the same minibatch yielded the best results.  
We overload the rotation operator to represent the PatchRot version of the image $\bmx_{pr} = R(s(\bmx); \iota_p)$, where we rotate all of $N$ patches of the input image. 
$\iota_p = \{\iota_1, \iota_2,..., \iota_N\}$ denotes the sequence of rotation applied to the corresponding input sequence $s(\bmx)$ and each $\iota_i\hspace{0.15cm} \forall \hspace{0.05cm} i=1,2,...,N$ is sampled randomly using a discrete uniform distribution. 

We train the ViT $T_\bmphi$ using $\bmx_r$ and $\bmx_{pr}$ images. 
The ViT $T_\bmphi = PE \odot EB \odot M$ consists of a patch encoding block $PE$ followed by encoder blocks $EB_i$ where $i\in\{1, 2, ..., e\}$ and e is the total number of encoder blocks, and an MLP head $M$ for classification. Let $E(s(\bmx))=\{E_0(s(\bmx)), E_1(s(\bmx)),...,E_N(s(\bmx))\}$ denote the output at the last encoder block $EB_e$ where $E_i(s(\bmx)) \in \mathbb{R}^{h}$ is the $i$-th output corresponding to the input sequence $s(\bmx)$ and $h$ is the embedding size. $E_0(s(\bmx))$ represents the encoding of the classification head and $\{E_1(s(\bmx)),$ $...,$ $E_N(s(\bmx))\}$ are the encoding for the patch heads. 
In the case of image rotation $\bmx_r$, the encoding from the classification head is passed to the MLP head $M_0$ to predict the rotation category of the image as $M_0(E_0(s(\bmx_r)))$. 
The encodings of the patch heads in this case are ignored. 
In the case of patch rotation, we introduce new MLPs $M_1,M_2,...,M_N$ to classify the encodings of the individual patches. 
The rotation angles of the individual patches are predicted by the MLPs as $M_i(E_i(s(\bmx_{pr}))) \hspace{0.15cm} \forall \hspace{0.05cm} i=1,2,...,N$.
PatchRot trains the ViT using,
\begin{equation}
    \sum_{j=0}^{3}\mathcal{L}_{ce}(M_0(E_0(s(R(\bmx, j)))), \bmy=j) + \sum_{i=1}^{N}\mathcal{L}_{ce}(M_i(E_i(R(s(\bmx), \iota_p))), \bmy=\theta_i),
\end{equation}

\noindent where the first term is the loss function penalizing rotation misclassification for each of the 4 angles of image rotation and the second term is the loss function penalizing rotation misclassification for the image patches.  $\mathcal{L}_{ce}$ is the standard cross-entropy loss function. 
Note that we do not rotate the image when rotating image patches as doing both simultaneously have demonstrated to hurt the downstream task performance.

\begin{figure*}[t]
    \begin{center}
      \includegraphics[width=\linewidth]{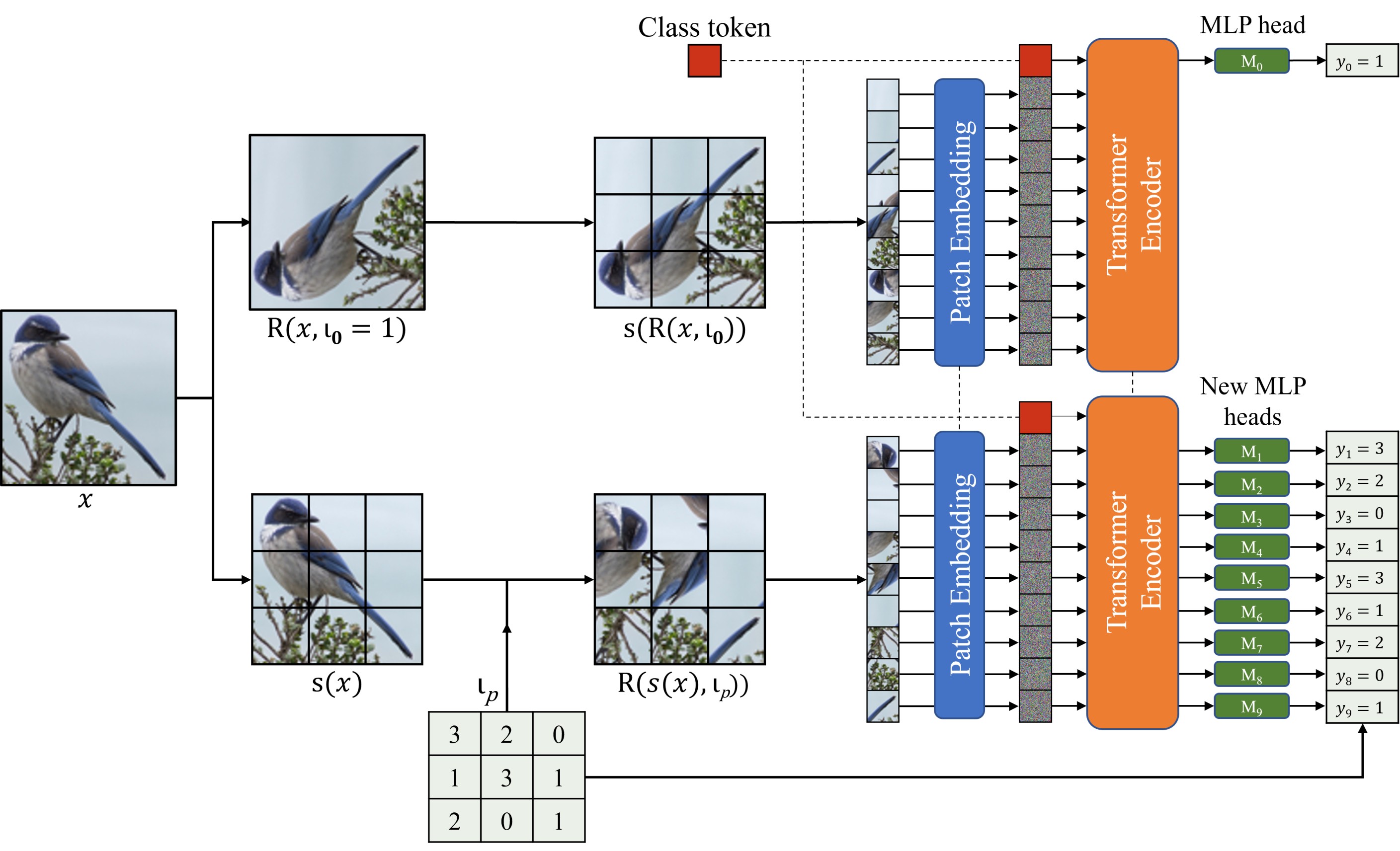}
    \end{center}
\caption{Model diagram of our approach. The input image is split into patches. The image and the patches are rotated by a random angle of $90$\textdegree$.\iota$ resulting in rotation by $0$\textdegree$, 90$\textdegree$, 180$\textdegree, or $270$\textdegree. The vision transformer (ViT) is trained to predict the rotation angle for the image and each patch using new additional MLP heads. Dotted lines denote the weights are shared. Best viewed in Color.}
\label{Fig:model_dig}
\end{figure*}

\subsection{Training Procedure}
\label{training_procedure}

To avoid the possibility of the network learning to predict the angle of patch rotations using edge continuity, we use a buffer gap $B$ between the patches, i.e., we initially partition the image using a larger patch size $P' = P+B$, where $P'>P$. 
Then from each such patch, we randomly crop a $P$ sized patch. 
This results in a gap between patches of random size between 0 to 2$B$ pixels. 
Due to the buffer between patches, the input size is reduced. Instead of scaling the image/patches to adjust to the original input size, we found it beneficial to perform the self-supervised training on reduced image size and use the original size for transferring knowledge to downstream tasks. 
To be precise, the PatchRot training image $\bmx_{pr}$ size is $C\times H_{pr}\times W_{pr}$ where $H_{pr} = P\cdot\lfloor\frac{H}{P+B}\rfloor$, and $W_{pr} = P\cdot\lfloor \frac{W}{P+B} \rfloor$ and $\lfloor.\rfloor$ denotes the floor operation. 
The number of patches is given by  $N_{pr}=\lfloor\frac{H}{P+B}\rfloor\cdot\lfloor \frac{W}{P+B} \rfloor$. 
For creating a rotated image $\bmx_{r}$, we produce a crop of this size instead of the original size for our algorithm and then rotate the cropped image.  
We were guided by the fact that training on smaller resolution images and then fine-tuning with higher resolution images has shown to yield performance gains (\cite{touvron2019fixing}).

Once the network is trained using PatchRot, we remove the newly added MLP heads - $M_1, M_2,...,M_N$ and only use the classification head's encoder output and its MLP head (just like the original ViT) for downstream tasks. 
For adapting the network to the new classification task, we replace the last classification layer from MLP head $M_{0}$ with a new layer having an output size equal to the number of categories in the new task before retraining the network. 
PatchRot training is performed at a smaller input size but for the downstream task, we use the original image size. The larger image size results in an increase in the number of input patches: $\lfloor\frac{H}{P+B}\rfloor\cdot\lfloor \frac{W}{P+B} \rfloor \rightarrow \frac{H}{P}\cdot\frac{W}{P}$. Hence, we apply a linear interpolation of the positional embedding as designed in the original ViT (\cite{dosovitskiy2020image}) for this case.

\begin{figure}[!ht]
     \centering
          \begin{subfigure}[b]{\textwidth}
         \centering
         \includegraphics[width=\textwidth]{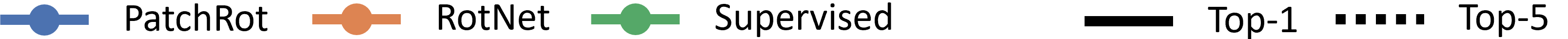}
         \label{fig:legend}
     \end{subfigure}

     \begin{subfigure}[b]{0.48\textwidth}
         \centering
         \includegraphics[width=\textwidth]{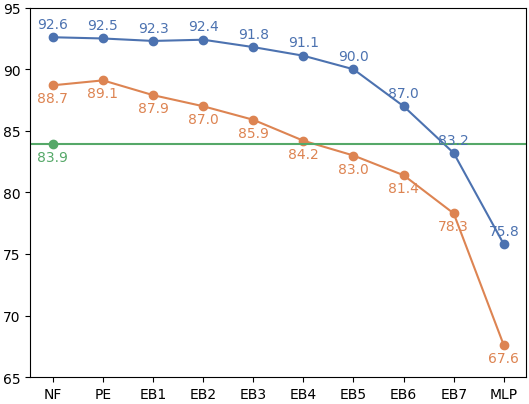}
         \caption{CIFAR-10}
         \label{c10_res}
     \end{subfigure}
    \hspace{0.25cm}
     \begin{subfigure}[b]{0.48\textwidth}
         \centering
         \includegraphics[width=\textwidth]{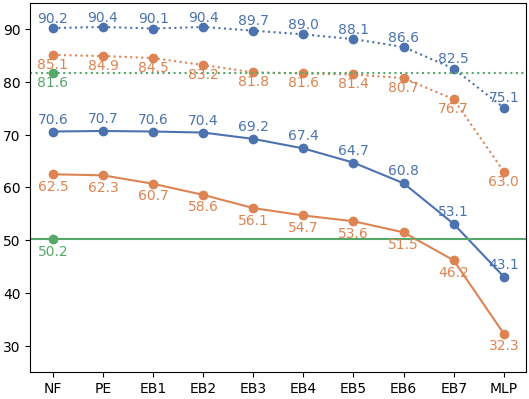}
         \caption{CIFAR-100}
         \label{c100_res}
     \end{subfigure}
     \begin{subfigure}[b]{0.48\textwidth}
         \centering
         \includegraphics[width=\textwidth]{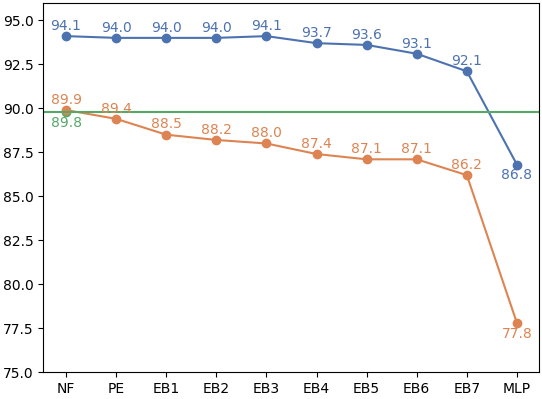}
         \caption{FashionMNIST}
         \label{fminst_res}
     \end{subfigure}
    \hspace{0.25cm}
     \begin{subfigure}[b]{0.48\textwidth}
         \centering
         \includegraphics[width=\textwidth]{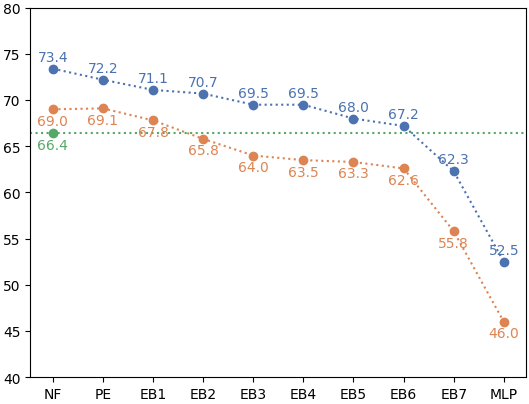}
         \caption{Tiny-ImageNet}
         \label{ti_res}
     \end{subfigure}
        \caption{Classification accuracies for (a) CIFAR-10, (b) CIFAR-100, (c) FashionMNIST and (d) Tiny-ImageNet datasets for \textcolor{green}{\textbf{Supervised}}, 
        \textcolor{Orange}{\textbf{RotNet}} (\cite{gidaris2018unsupervised})
        and \textcolor{blue}{\textbf{PatchRot}}.
        Solid lines denote Top-1 accuracy and the dashed lines denote Top-5 accuracy on the Y-axis. The X-axis contains different layers of the transformer. While finetuning on the downstream task all the layers before that layer are frozen.
        NF: No layers are frozen; 
        PE: Patch Embedding block;
        EB-1 to EB-7 are the encoder blocks of the Vision Transformer.
        MLP: The whole network is frozen except for the new output linear layer.}
        \label{fig:results}
\end{figure}

\section{Experiments}
\subsection{Datasets}
We test our approach on datasets like CIFAR-10, CIFAR-100, FashionMNIST, and Tiny-ImageNet datasets using a scaled-down version of the ViT (described in section \ref{trainingdetails}) due to hardware limitations. 
We use $32 \times 32$ image size for CIFAR-10, CIFAR-100 and FashionMNIST datasets and $64 \times 64$ for Tiny-ImageNet. Random crop with zero padding of size 4 and a random horizontal flip is used as augmentations for CIFAR-10, CIFAR-100, and Tiny-ImageNet datasets. We did not apply any augmentations for FashionMNIST. Also, we do not use any padding while creating patch rotation image $\bmx_{pr}$ as padding pixels tend to work as a shortcut to find the rotation angle.

\subsection{Training Details}
\label{trainingdetails}
The experiments are performed using ViT-Lite from (\cite{hassani2021escaping}) (scaled-down version of the original ViT architecture (\cite{dosovitskiy2020image})  which is designed for smaller datasets. Specifically, the number of encoder blocks is reduced to 6 with 256 embedding size and 512 expansion size. The number of attention heads is also reduced to 4 and the dropout is set to 0.1. We train the network with the patch size of $P=4$ for CIFAR-10, CIFAR-100, and FashionMNIST and $P=8$ for TinyImageNet. The buffer $B$ is set to $\frac{1}{4}$ of the patch size which results in PatchRot input size of $24\times24$  for $32\times32$ images and $48\times48$ for $64\times64$ images.

We used Adam Optimizer for training the ViT with a learning rate and a weight decay of $5\times 10^{-4}$ and $3\times 10^{-2}$ respectively.
The learning rate is warmed up for the first 10 epochs and then decayed using a cosine schedule.
The batch size is set to 128. 
However, due to multiple variations of the same image being presented in a mini-batch, the effective batch size for PatchRot is 128$\times$5 samples. 

We follow the procedure of fine-tuning the layers after the self-supervision training as also implemented by (\cite{noroozi2016unsupervised}) for the experiments.
The self-supervised training is performed for 300 epochs and the supervised training is performed for 200 epochs. We provide additional experimental results with different input patch sizes in the appendix. The code for PatchRot is available at: \url{https://github.com/s-chh/PatchRot}.

\subsection{Results}
We compare our approach with supervised learning and image rotation prediction (\cite{gidaris2018unsupervised}). RotNet is a convolutional neural network approach. We used the code from their offical repo and used it to train the ViT with the same setting as PatchRot.
The results are in Figure \ref{fig:results}, and the tabular versions are available in the appendix.
In Figure \ref{fig:results}, NF denotes when no layers are frozen and the whole network was trained. MLP denotes linear probing of the learned features where we freeze the whole network and only train the last classification layer. The middle results denote results on fine-tuning specific encoder blocks.
Fine-tuning just the MLP head (one fully connected layer) achieves results close to the supervised learning and fine-tuning one encoder block and MLP head outperforms the supervised training from scratch.

\section{Analysis}

\begin{table*}
\newcommand{\columnlen}{0.8cm}
\begin{center}
\begin{tabular}{l|c   c  cc  c  cc  c  c c|}
\hline
\toprule
Initalization & NF & PE & EB1 & EB2 & EB3 & EB4 & EB5 & EB6 & EB7 & MLP\\
\midrule
No ImageRot & 91.8 & 91.9 & 91.9 & 91.4 & 91.0 & 90.0 & 88.8 & 82.2 & 69.4 & 54.9\\
No PatchRot & 91.0 & 91.2 & 90.7 & 90.2 & 89.4 & 88.6 & 87.8 & 85.3 & 80.1 & 70.8\\
Original Size & 92.1 & 92.2 & 91.6 & 91.1 & 90.7 & 89.9 & 89.0 & 86.2 & 81.6 & 73.9\\
Rotate Img \& Patch&90.7 & 90.7 & 90.8 & 90.6 & 90.6 & 90.2 & 88.3 & 82.5 & 72.8 & 58.4\\
Reuse MLP head & 91.1 & 91.0 & 90.9 & 90.4 & 89.9 & 89.6 & 87.7 & 84.0 & 76.6 & 65.7\\
\midrule
PatchRot-full & \textbf{92.6} & \textbf{92.5} & \textbf{92.3} & \textbf{92.4} & \textbf{91.8} & \textbf{91.1} & \textbf{90.0} & \textbf{87.0} & \textbf{83.2} & \textbf{75.8}\\
\bottomrule
\end{tabular}
\end{center}
\caption{Ablation study of our method on CIFAR-10. 
Columns denote the parts of the ViT being fine-tuned and all the layers before that layer are frozen. 
NF: No layers are frozen; 
PE: Patch Embedding block;
EB-1 to EB-7 are the encoder blocks of the Vision Transformer.
MLP: The whole network is frozen except for the new output linear layer.
}
\label{c10_ablation}
\end{table*}

\begin{figure}[t]
     \centering
     \begin{subfigure}[b]{0.5\textwidth}
         \centering
         \includegraphics[width=\textwidth]{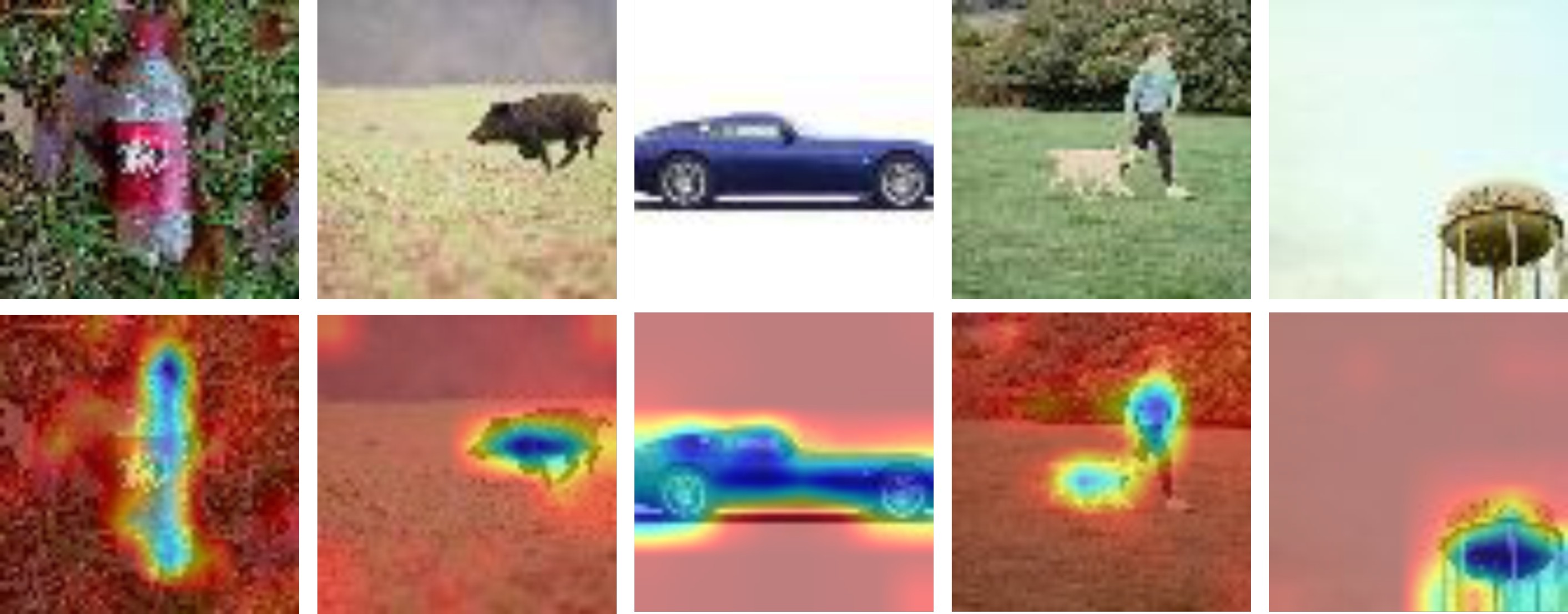}
         \caption{Attention Maps}
         \label{att1}
     \end{subfigure}
    \hspace{0.5cm}
     \begin{subfigure}[b]{0.4\textwidth}
         \centering
         \includegraphics[width=\textwidth]{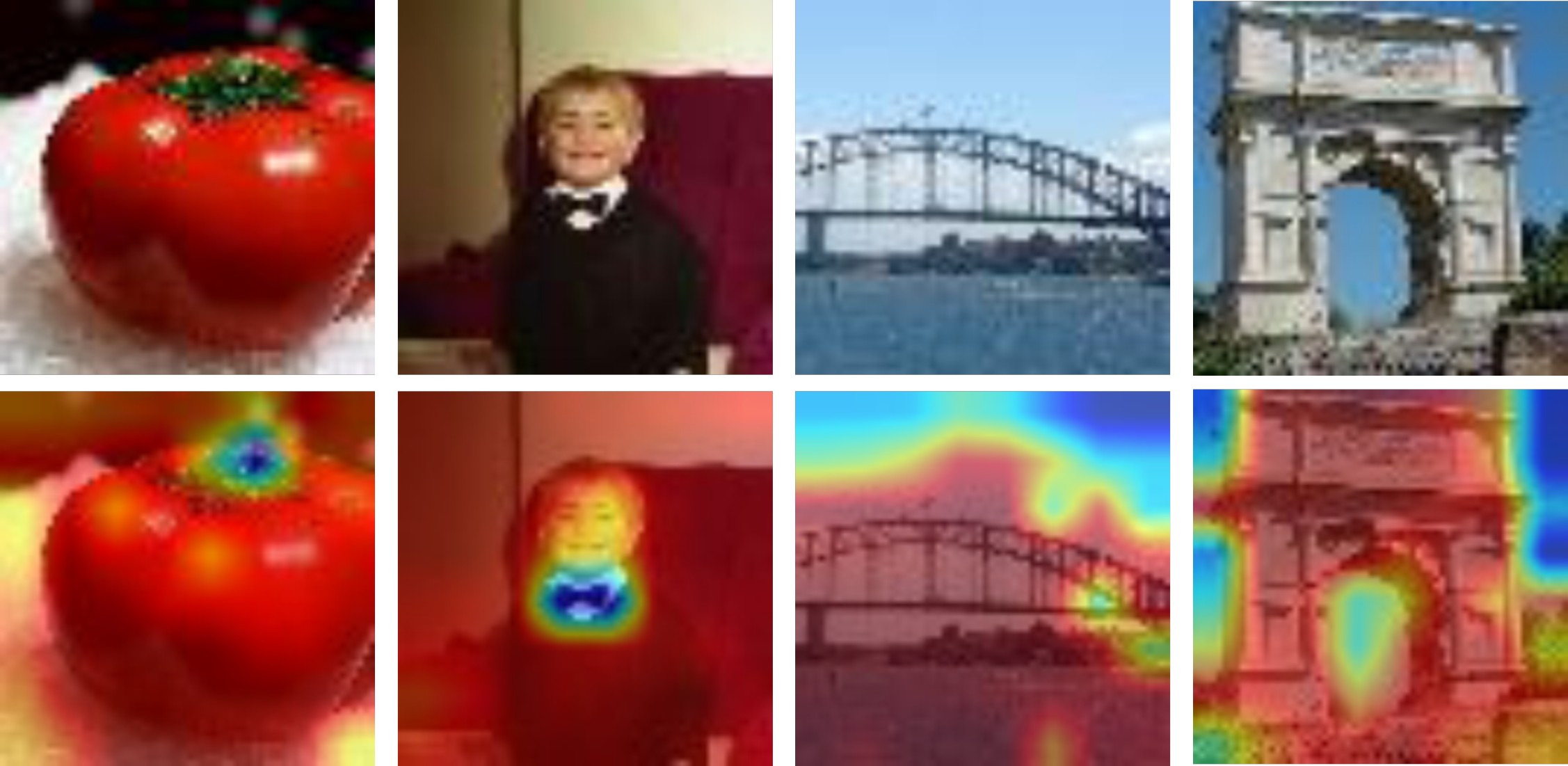}
         \caption{Failure cases}
         \label{att2}
     \end{subfigure}
     \caption{Sample Attention Maps of ViT trained using PatchRot on the validation set of Tiny-ImageNet. Upper row: Input and Lower row: Attention Map}
     \label{Fig:attmap}
\end{figure}

\subsection{Ablation Study}
In this section, we test the importance of different components of PatchRot using CIFAR-10 and present the results in Table \ref{c10_ablation}.
First, we train the ViT on the patch rotation image $\bmx_{pr}$ only and exclude the image rotation $\bmx_{r}$ versions (No ImageRot). 
As we do not train the classification head, the network learns only the local characteristics and as we freeze more layers, we can notice the drop in performance, signifying the importance of global context. 
Similarly, in the second experiment, we exclude the $\bmx_{pr}$ (No PatchRot). 
This is comparable to the RotNet which is trained to predict image rotation using a convolutional network with the only difference that the self-supervised training is performed at a reduced image size.
Next, we test the significance of reduced size training by comparing it with self-supervised training on the original image scale (Original Size). 
To test this, we divide the image using the original patch size $P$ and we randomly crop $P-B$ sized patches from it. These cropped patches are then resized to patch size $P$ to maintain the original image size and still have a buffer.
We also test our hypothesis of rotating images and patches together (Section \ref{method}) in our next experiment (Rotate Img \& Patch). 
Last, we experiment with the newly added MLP heads (Reuse MLP Head). 
Since adding new MLP heads temporarily increases the model size. Instead of adding new MLP heads, we use the same original MLP head for predicting the rotation angle for the image and the patches.

\subsection{Attention Maps}
We show the attention maps on the validation set of Tiny-ImageNet for ViT trained using PatchRot in Figure \ref{Fig:attmap}. With just the unsupervised training, the model learns to attend to the main object in the image. We also show failure cases in Figure \ref{att2} where the model learns to solve the problem by attending to a part of the object like the top green part of the tomato or using a background feature like sky positioning. More attention maps are available in the appendix.

\begin{figure*}[t]
    \raggedright
      \includegraphics[scale=0.15]{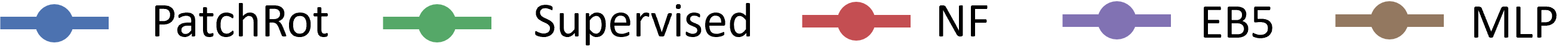}
    \begin{center}
      \includegraphics[width=\textwidth]{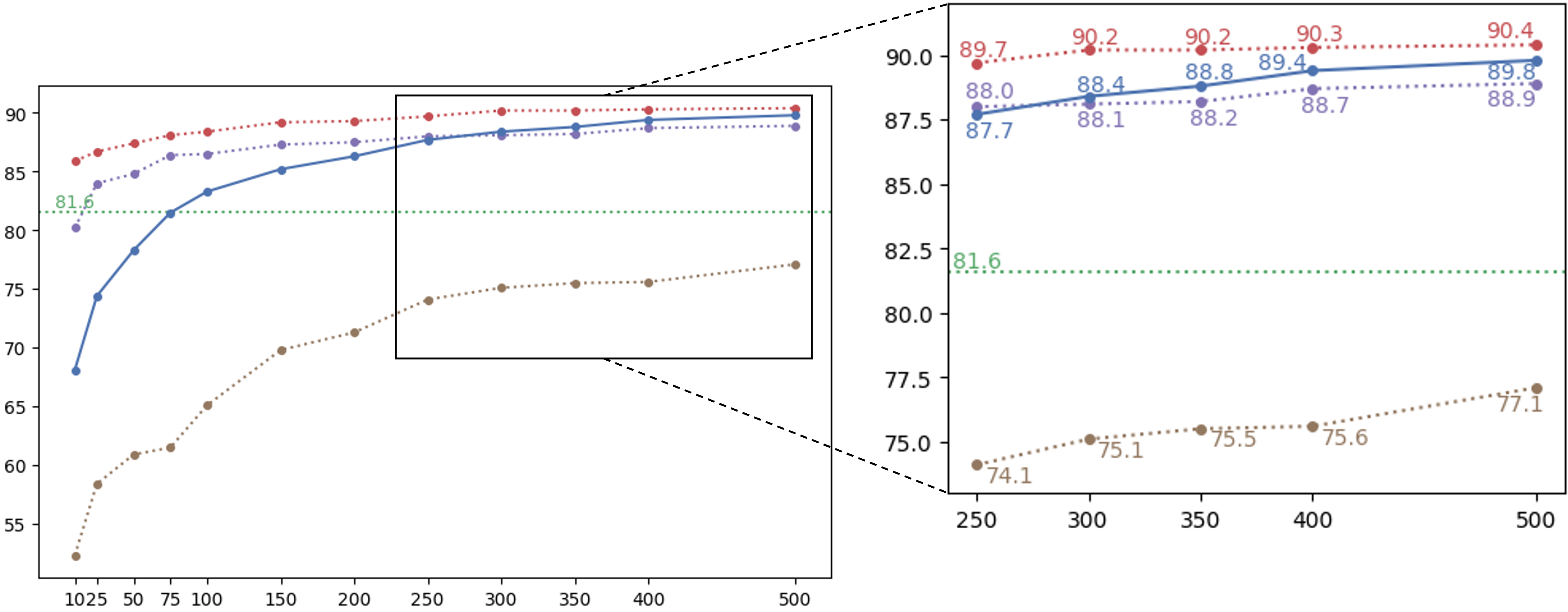}
    \end{center}
\caption{Top-5 classification accuracy vs Pretraining epochs for CIFAR-100. \textcolor{green}{Supervised} denotes the supervised accuracy; \textcolor{blue}{PatchRot} denotes the PatchRot self-supervised accuracy on testset; \textcolor{red}{NF}, \textcolor{purple}{EB5} and \textcolor{brown}{MLP} denotes the accuracy on finetuning all the layers, freezing layers till EB5 and no fine-tuning of the network respectively. Best viewed in color.}
\label{Fig:pretrain}
\end{figure*}

\subsection{Pre-training epochs vs Classification Accuracy}
Here, we study the impact of PatchRot's pretraining on the final classification performance. We train the ViT using our approach with a different number of epochs \{10, 25, 50, 75, 100, 150, 200, 250, 300, 350, and 400\}, for the CIFAR-100 dataset. 
The results are in Figure \ref{Fig:pretrain}. Pretraining with just a few epochs with PatchRot we can observe significant improvement in final performance compared to training from scratch (Supervised). 
Also, PatchRot did not show any signs of overfitting, and training it longer (350, 400, and 500) improves the self-supervised test accuracy and the final classification accuracy.

\begin{figure}[t]
     \centering
      \begin{subfigure}[b]{\textwidth}
         \centering
         \includegraphics[width=\textwidth]{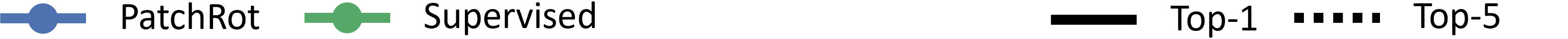}
         \label{fig:xlegend}
     \end{subfigure}
     \begin{subfigure}[b]{0.48\textwidth}
         \centering
         \includegraphics[width=\textwidth]{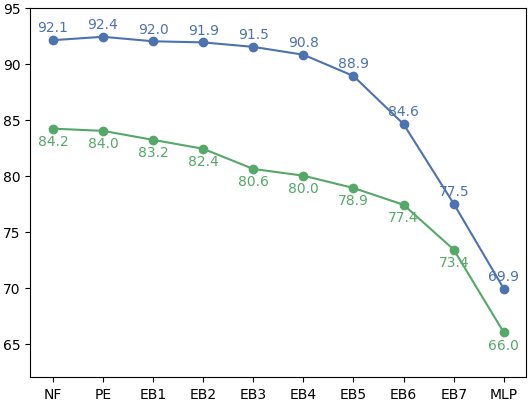}
         \caption{CIFAR100$\rightarrow$CIFAR10}
         \label{xc10}
     \end{subfigure}
    \hspace{0.25cm}
     \begin{subfigure}[b]{0.48\textwidth}
         \centering
         \includegraphics[width=\textwidth]{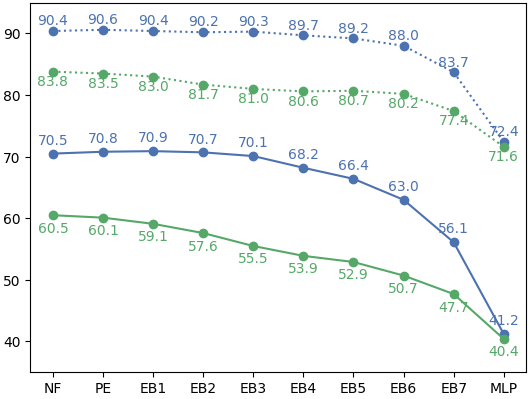}
         \caption{CIFAR10$\rightarrow$CIFAR100}
         \label{xc100}
     \end{subfigure}
     \caption{PatchRot performance in a transfer learning setting. \textcolor{blue}{PatchRot} denotes the ViT was trained on the source dataset self-supervised using PatchRot. \textcolor{green}{Supervised} denotes the ViT was trained on source using the supervised object classification task.
     Solid lines and the dashed lines show the Top-1 and Top-5 accuracy on the Y-axis respectively.
     The X-axis contains different layers of the transformer and while finetuning on the target dataset all the layers before that layer are frozen.
     }
     \label{Fig:transfer}
\end{figure}

\subsection{Transfer Learning}
PatchRot aims to learn rich features in an unsupervised manner and in this section, we show its advantage for the transfer learning settings on CIFAR100 $\rightarrow$ CIFAR10 and CIFAR10 $\rightarrow$ CIFAR100 tasks. 
We first train the network on source data using PatchRot vs. the supervised object classification task. 
These networks are used as the initialization to train on the target dataset and the results are in Figure \ref{Fig:transfer}. 
We can observe that PatchRot extracts better features than the supervised training which can be used across datasets. Also, PatchRot has a significant margin advantage over supervised training which drops only during the last few layers (where layers become domain/task-specific). 
Hence, pre-training the network with PatchRot has a significant advantage over supervised training.

\begin{table*}[t]
\newcommand{\columnlen}{0.8cm}
\begin{center}
\begin{tabular}{c|c |  c  cc  c  cc  c  c c c|}
\hline
\toprule
Labels & Sup & NF & PE & EB1 & EB2 & EB3 & EB4 & EB5 & EB6 & EB7 & MLP\\
\midrule
40 & 20.8 & 38.8 & 40.1 & 42.1 & 44.5 & 46.9 & 49.7 & 50.6 & 51.3 & \textbf{55.2} & 27.5\\
250 & 26.6 & 57.5 & 60.2 & 61.4 & 63.4 & 64.9 & 65.3 & \textbf{65.8} & 65.1 & 65.2 & 35.5\\
1000 & 38.6 & 70.1 & 71.4 & 71.9 & 72.6 & 72.8 & \textbf{73.8} & 73.5 & 72.0 & 72.2 & 58.4\\
4000 & 53.7 & 79.2 & 79.6 & 80.7 & 80.9 & \textbf{81.1} & 80.9  & 80.2 & 78.4 & 77.2 & 69.7\\
10000 & 67.1 & 84.7 & 85.2 & 85.5 & \textbf{85.6} & 85.5 & 84.8 & 84.3 & 82.2 & 79.6 & 73.0\\
20000& 75.2 & 88.3 & 88.7 & \textbf{88.9} & \textbf{88.9} & 88.5 & 88.0 & 86.7 & 84.7 & 81.1 & 74.5\\
30000& 79.8 & 90.0 & 90.5 & \textbf{90.6} & 90.3 & 89.9 & 89.2 & 88.1 & 85.9 & 82.0 & 75.0\\
40000& 82.0 & 91.7 & \textbf{92.0} & 91.7 & 91.6 & 91.1 & 90.3 & 89.2 & 86.4 & 82.8 & 75.6\\
\midrule
50000 & 83.9 & \textbf{92.6} & 92.5 & 92.3 & 92.4 & 91.8 & 91.1 & 90.0 & 87.0 & 83.2 & 75.8\\
\bottomrule
\end{tabular}
\end{center}
\caption{Results on CIFAR10 in a semi-supervised setting. PatchRot training is performed on all the samples and then the network is fine-tuned with labeled samples only. The first column denotes the number of labeled samples and the second column shows the supervised performance with labeled samples only. Other columns denote the parts of the ViT being fine-tuned and all the layers before that layer are frozen. 
}
\label{c10_ssl}
\end{table*}

\subsection{Application to Semi-supervised learning}
Self-supervised learning techniques are being popularly utilized for semi-supervised learning by training the network with labeled data and simultaneously training another output head of the network with the unlabeled data using the self-supervised loss (\cite{gidaris2018unsupervised,zhai2019s4l}). 
In our experiments, instead of training the ViT on both the supervised and PatchRot loss together, we first train the network with PatchRot (self-supervised loss) on all the data and then fine-tune the network using the labeled data only (Similar to previous experiments). We use CIFAR-10 for these experiments with \{40, 250, 1000, 4000, 10000, 20000, 30000 and 40000\} labeled samples and the results are in Table \ref{c10_ssl}. The second column (Sup) denotes the test accuracy on training the network with labeled samples only. As expected, with the increase in the number of labeled samples, training more layers is beneficial. PatchRot pre-training results in superior performance than the Supervised training showcasing its application in semi-supervised learning.


\begin{figure}[!ht]
     \centering
          \begin{subfigure}[b]{\textwidth}
         \centering
         \includegraphics[width=\textwidth]{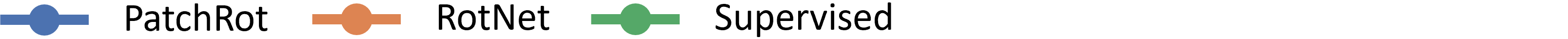}
         \label{fig:legend2}
     \end{subfigure}
     \begin{subfigure}[b]{0.48\textwidth}
         \centering
         \includegraphics[width=\textwidth]{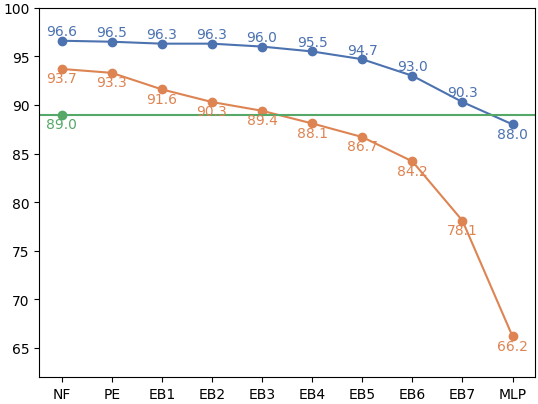}
         \caption{SVHN}
         \label{svhn_res}
     \end{subfigure}
    \hspace{0.25cm}
     \begin{subfigure}[b]{0.48\textwidth}
         \centering
         \includegraphics[width=\textwidth]{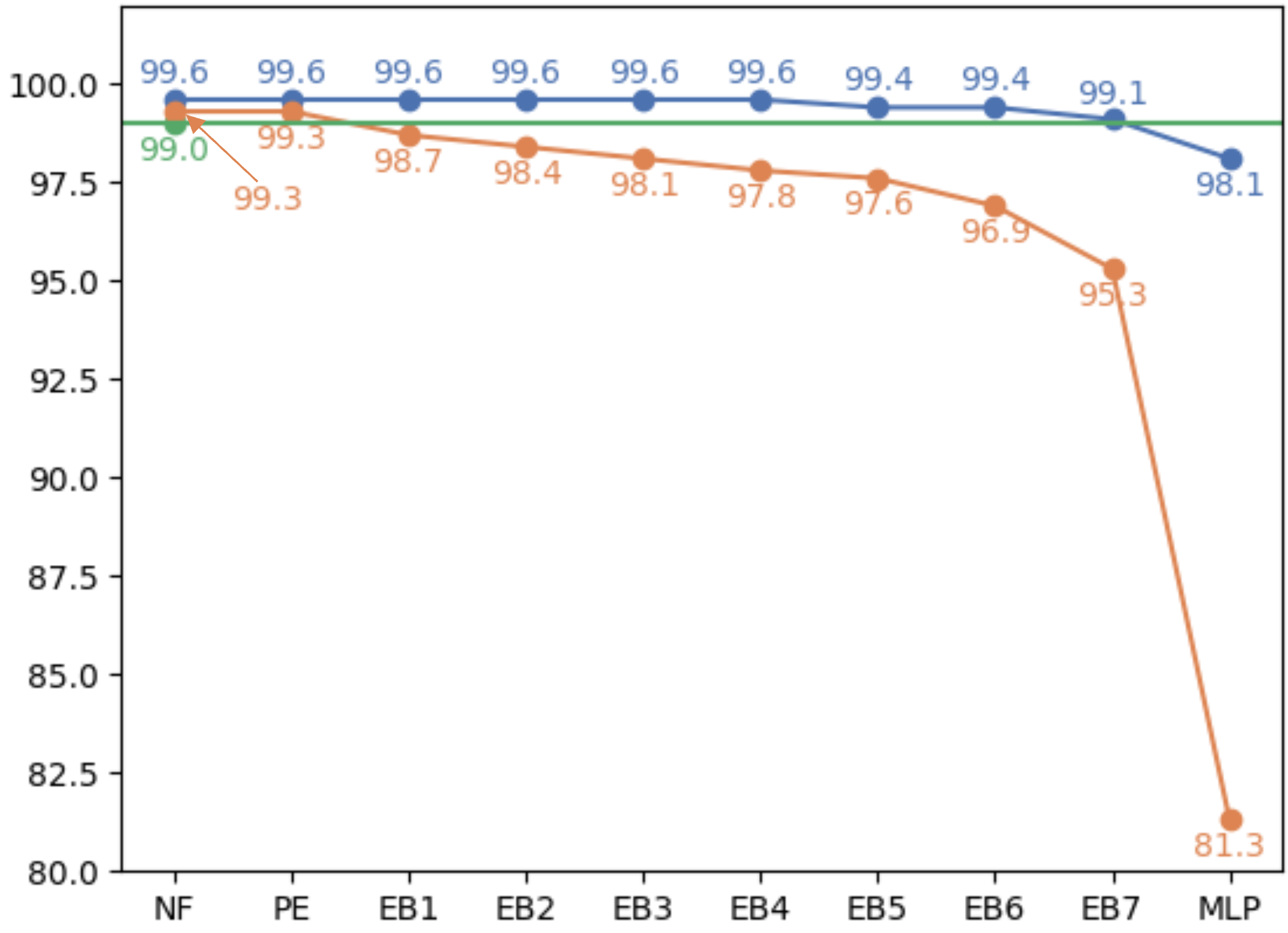}
         \caption{MNIST}
         \label{mniist_res}
     \end{subfigure}
        \caption{Top-1 classification accuracies for Rotation Invariant datasets: (a) SVHN and (b) MNIST for \textcolor{green}{\textbf{Supervised}}, 
        \textcolor{Orange}{\textbf{RotNet}} (\cite{gidaris2018unsupervised})
        and \textcolor{blue}{\textbf{PatchRot}}.
        Solid lines denote Top-1 accuracy and the dashed lines denote Top-5 accuracy on the Y-axis. The X-axis contains different layers of the transformer. While finetuning on the downstream task all the layers before that layer are frozen.
        }
        \label{fig:svhn_4}
\end{figure}

\subsection{The Case of Rotation Invariant Objects}
Predicting the rotation angle of an object works only if the object is distinguishable when rotated by different angles. PatchRot also performs rotation on the image patches and predicting their rotation angle still requires knowledge of the object in the image. 
However, some patches can be rotation invaritant too but those patches are generally part of the background. Hence, PatchRot is helpful for rotation invariant objects as well. We perform this experiment on the digits datasets: SVHN and MNIST where digits like 0, 1, and 8 are rotation invariant whereas the other digits can differentiate the rotation angles. We use the same experiment setting as the other datasets and the results are shown in Figure \ref{fig:svhn_4}. RotNet (\cite{gidaris2018unsupervised}) provides superior performance than supervised due to some rotation discriminative digits (like 2, 3, 4, etc) but PatchRot gets a significantly higher performance than Image-level rotation prediction and supervised training.

\section{Conclusion}
In this paper, we proposed PatchRot which is an easy-to-understand and implement self-supervised technique. PatchRot trains a Vision transformer to predict rotation angles $\in \{$0 \textdegree, $90$\textdegree, $180$\textdegree, $270$\textdegree \} of image and image patches. With extensive experiments on multiple datasets, we showcased that a Vision transformer pretrained with our approach achieves superior results on downstream supervised learning. PatchRot is a robust technique that works for rotation invariant objects as well and can be applied in transfer and semi-supervised learning settings. 

\begin{ack}
The work was supported in part by a grant from ONR. Any opinions expressed in this material are those of the authors and do not necessarily reflect the views of ONR.
\end{ack}

{\small
\bibliographystyle{abbrvnat}
\bibliography{egbib}
}

\clearpage
\appendix

\section{Appendix}

\subsection{Tablular and Additional Results}
In the main paper, we provided the comparison against RotNet and supervised training in graphs. Here, we provide the results in tabular format for easy lookup. We also experiment with training ViT using different patch sizes and compare the approach with random initialization in a similar manner. The results for these are in Table \ref{c10_4_tab_res},\ref{c10_8_tab_res},\ref{c100_4_tab_res},\ref{c100_8_tab_res},\ref{fmnist_4_tab_res},\ref{fmnist_8_tab_res},\ref{ti_8_tab_res} and \ref{ti_16_tab_res}.

\begin{table*}[hb]
\newcommand{\columnlen}{0.8cm}
\begin{center}
\begin{tabular}{l|c  c  cc  c  cc  c  c  c|}
\toprule
Initalization & NF & PE & EB1 & EB2 & EB3 & EB4 & EB5 & EB6 & EB7 & MLP\\
\midrule
Random & 83.9 & 77.2 & 75.8 & 75.6 & 74.0 & 70.9 & 65.6 & 52.9 & 41.9 & 34.3\\
RotNet  & 88.7 & 89.1 & 87.9 & 87.0 & 85.9 & 84.2 & 83.0 & 81.4 & 78.3 & 67.6\\
PatchRot (Ours) & \textbf{92.6} & \textbf{92.5} & \textbf{92.3} & \textbf{92.4} & \textbf{91.8} & \textbf{91.1} & \textbf{90.0} & \textbf{87.0} & \textbf{83.2} & \textbf{75.8}\\
\bottomrule
\end{tabular}
\end{center}
\caption{Top-1 Classification accuracies on CIFAR-10  using patch size of 4. }
\label{c10_4_tab_res}
\end{table*}

\begin{table*}[hb]
\newcommand{\columnlen}{0.8cm}
\begin{center}
\begin{tabular}{l|c c  cc  c  cc  c  c c|}
\toprule
Initalization & NF & PE & EB1 & EB2 & EB3 & EB4 & EB5 & EB6 & EB7 & MLP\\
\midrule
Random & 79.0 & 73.6 & 72.8 & 72.2 & 71.8 & 70.2 & 65.8 & 50.7 & 39.3 & 31.8\\
RotNet  & 85.3 & 85.5 & 84.5 & 83.4 & 82.7 & 81.5 & 80.2 & 78.2 & 74.9 & 66.1\\
PatchRot (Ours) & \textbf{87.2} & \textbf{87.0} & \textbf{86.6} & \textbf{86.2} & \textbf{84.9} & \textbf{84.3} & \textbf{83.0} & \textbf{80.2} & \textbf{77.0} & \textbf{70.0}\\
\bottomrule
\end{tabular}
\end{center}
\caption{Top-1 Classification accuracies on CIFAR-10 using patch size of 8. }
\label{c10_8_tab_res}
\end{table*}

\begin{table*}[h]
\footnotesize
\newcommand{\columnlen}{0.8cm}
\begin{center}
\begin{tabular}{l| c|c c  cc  c  cc  c  c c|}
\toprule
Initalization & Metric & NF & PE & EB1 & EB2 & EB3 & EB4 & EB5 & EB6 & EB7 & MLP\\
\midrule
\multirow{2}{*}{Random} & Top-1 & 50.2 & 45.4 & 44.3 & 43.3 & 41.9 & 40.4 & 37.7 & 25.3 & 14.8 & 10.1
\\
 & Top-5 & 81.6 & 74.2 & 74.0 & 73.2 & 72.7 & 70.5 & 66.7 & 55.2 & 42.8 & 33.8
\\
\midrule
\multirow{2}{*}{RotNet } & Top-1 & 62.5 & 62.3 & 60.7 & 58.6 & 56.1 & 54.7 & 53.6 & 51.5 & 46.2 & 32.3\\
& Top-5 & 85.1 & 84.9 & 84.5 & 83.2 & 81.8 & 81.6 & 81.4 & 80.7 & 76.7 & 63.0\\
\midrule
\multirow{2}{*}{PatchRot} & Top-1 & \textbf{70.6} & \textbf{70.7} & \textbf{70.6} & \textbf{70.4} & \textbf{69.2} & \textbf{67.4} & \textbf{64.7} & \textbf{60.8} & \textbf{53.1} & \textbf{43.1}\\
& Top-5 & \textbf{90.2} & \textbf{90.4} & \textbf{90.1} & \textbf{90.4} & \textbf{89.7} & \textbf{89.0} & \textbf{88.1} & \textbf{86.6} & \textbf{82.5} & \textbf{75.1}\\
\bottomrule
\end{tabular}
\end{center}
\caption{Top-1 and Top-5 Classification accuracies on CIFAR-100  using patch size of 4.}
\label{c100_4_tab_res}
\end{table*}

\begin{table*}[t]
\footnotesize
\newcommand{\columnlen}{0.8cm}
\begin{center}
\begin{tabular}{l| c|c c  cc  c  cc  c  c c|}
\toprule
Initalization & Metric & NF & PE & EB1 & EB2 & EB3 & EB4 & EB5 & EB6 & EB7 & MLP\\
\midrule
\multirow{2}{*}{Random} & Top-1 & 50.2 & 45.4 & 44.3 & 43.3 & 41.9 & 40.4 & 37.7 & 25.3 & 14.8 & 10.1\\
 & Top-5 & 76.5 & 71.1 & 70.6 & 69.6 & 69.2 & 67.9 & 65.8 & 52.5 & 38.2 & 29.8\\
\midrule
\multirow{2}{*}{RotNet }& Top-1 & 54.1 & 53.5 & 51.6 & 50.4 & 47.9 & 47.5 & 46.2 & 45.9 & 40.6 & 29.4\\
& Top-5 & 80.3 & 80.2 & 79.0 & 77.9 & 77.1 & 76.0 & 75.7 & 75.5 & 71.1 & 59.4\\
\midrule
\multirow{2}{*}{PatchRot} & Top-1 &\textbf{60.5} & \textbf{61.2} & \textbf{60.9} & \textbf{58.7} & \textbf{57.1} & \textbf{54.5} & \textbf{53.8} & \textbf{52.2} & \textbf{45.2} & \textbf{34.0}\\
& Top-5 & \textbf{84.9} & \textbf{85.6} & \textbf{85.1} & \textbf{84.2} & \textbf{83.3} & \textbf{82.4} & \textbf{81.6} & \textbf{81.1} & \textbf{75.6} & \textbf{66.0}\\
\bottomrule
\end{tabular}
\end{center}
\caption{Top-1 and Top-5 Classification accuracies on CIFAR-100  using patch size of 8.}
\label{c100_8_tab_res}
\end{table*}

\begin{table*}
\newcommand{\columnlen}{0.8cm}
\begin{center}
\begin{tabular}{l|c c  cc  c  cc  c  c c|}
\toprule
Initalization & NF & PE & EB1 & EB2 & EB3 & EB4 & EB5 & EB6 & EB7 & MLP\\
\midrule
Random & 89.8 & 87.8 & 88.2 & 88.2 & 87.9 & 87.4 & 84.5 & 79.9 & 74.9 & 63.1\\
RotNet  & 89.9 & 89.4 & 88.5 & 88.2 & 88.0 & 87.4 & 87.1 & 87.1 & 86.2 & 77.8\\
PatchRot (Ours) & \textbf{94.1} & \textbf{94.0} & \textbf{94.0} & \textbf{94.0} & \textbf{94.1} & \textbf{93.7} & \textbf{93.6} & \textbf{93.1} & \textbf{92.1} & \textbf{86.8}\\
\bottomrule
\end{tabular}
\end{center}
\caption{Top-1 Classification accuracies on FashionMNIST using patch size of 4.}
\label{fmnist_4_tab_res}
\end{table*}

\begin{table*}
\newcommand{\columnlen}{0.8cm}
\begin{center}
\begin{tabular}{l|c c  cc  c  cc  c  c c|}
\toprule
Initalization & NF & PE & EB1 & EB2 & EB3 & EB4 & EB5 & EB6 & EB7 & MLP\\
\midrule
Random & 90.7 & 76.5 & 89.6 & 89.1 & 88.9 & 88.7 & 87.9 & 85.6 & 80.9 & 71.8\\
RotNet  & 91.1 & 90.9 & 90.4 & 90.2 & 89.2 & 89.0 & 88.9 & 87.9 & 87.1 & 76.7\\
PatchRot (Ours) & \textbf{92.0} & \textbf{92.4} & \textbf{92.7} & \textbf{92.8} & \textbf{92.6} & \textbf{92.3} & \textbf{92.4} & \textbf{91.4} & \textbf{89.8} & \textbf{85.2}\\
\bottomrule
\end{tabular}
\end{center}
\caption{Top-1 Classification accuracies on FashionMNIST using patch size of 8.}
\label{fmnist_8_tab_res}
\end{table*}

\begin{table*}
\footnotesize
\newcommand{\columnlen}{0.8cm}
\begin{center}
\begin{tabular}{l| c|c c  cc  c  cc  c  c c|}
\toprule
Initalization & Metric & NF & PE & EB1 & EB2 & EB3 & EB4 & EB5 & EB6 & EB7 & MLP\\
\midrule
\multirow{2}{*}{Random} & Top-1 & 41.9 & 36.6 & 35.4 & 33.5 & 33.2 & 31.9 & 29.4 & 18.9 & 9.9 & 6.9\\
& Top-5 & 66.4 & 61.4 & 60.1 & 59.5 & 59.3 & 57.7 & 54.4 & 39.5 & 26.4 & 19.7\\
\midrule
\multirow{2}{*}{RotNet }& Top-1 & 45.1 & 45.2 & 43.8 & 41.0 & 38.4 & 36.9 & 36.4 & 36.3 & 30.7 & 22.2\\
& Top-5 & 69.0 & 69.1 & 67.8 & 65.8 & 64.0 & 63.5 & 63.3 & 62.6 & 55.8 & 46.0\\
\midrule
\multirow{2}{*}{PatchRot} & Top-1 & \textbf{48.6} & \textbf{47.4} & \textbf{45.5} & \textbf{44.6} & \textbf{43.2} & \textbf{42.7} & 41.6 & 40.0 & 35.5 & 26.3\\
& Top-5 & \textbf{73.4} & \textbf{72.2} & \textbf{71.1} & \textbf{70.7} & \textbf{69.5} & \textbf{69.5} & \textbf{68.0} & \textbf{67.2} & \textbf{62.3} & \textbf{52.5}\\
\bottomrule
\end{tabular}
\end{center}
\caption{Top-1 and Top-5 Classification accuracies on Tiny-ImageNet using patch size of 8.}
\label{ti_8_tab_res}
\end{table*}

\begin{table*}
\footnotesize
\newcommand{\columnlen}{0.8cm}
\begin{center}
\begin{tabular}{l| c|c c  cc  c  cc  c  c c|}
\toprule
Initalization & Metric & NF & PE & EB1 & EB2 & EB3 & EB4 & EB5 & EB6 & EB7 & MLP\\
\midrule
\multirow{2}{*}{Random} & Top-1 & 33.9 & 29.9 & 29.5 & 28.5 & 27.7 & 27.4 & 25.8 & 15.2 & 8.3 & 5.6\\
& Top-5 & 59.3 & 54.7 & 54.1 & 53.1 & 52.5 & 52.3 & 49.9 & 34.6 & 23.1 & 17.3\\
\midrule
\multirow{2}{*}{RotNet }& Top-1 & 35.8 & 35.6 & 33.7 & 31.5 & 30.8 & 30.2 & 29.4 & 29.6 & 27.6 & 22.2\\
& Top-5 & 61.8 & 61.1 & 59.5 & 57.8 & 56.7 & 56.3 & 55.8 & 56.1 & 53.4 & 46.6\\
\midrule
\multirow{2}{*}{PatchRot} & Top-1 & \textbf{38.2} & \textbf{37.8} & \textbf{37.0} & \textbf{36.2} & \textbf{34.7} & \textbf{34.4} & \textbf{33.9} & \textbf{33.1} & \textbf{29.1} & \textbf{22.8}\\
& Top-5 & \textbf{65.1} & \textbf{64.7} & \textbf{63.8} & \textbf{62.9} & \textbf{61.7} & \textbf{61.5} & \textbf{60.7} & \textbf{59.6} & \textbf{56.1} & \textbf{46.8}\\
\bottomrule
\end{tabular}
\end{center}
\caption{Top-1 and Top-5 Classification accuracies on Tiny-ImageNet using patch size of 16.}
\label{ti_16_tab_res}
\end{table*}

\subsection{More Attention Maps}
\begin{figure}[t]
    \begin{center}
      \includegraphics[width=\linewidth]{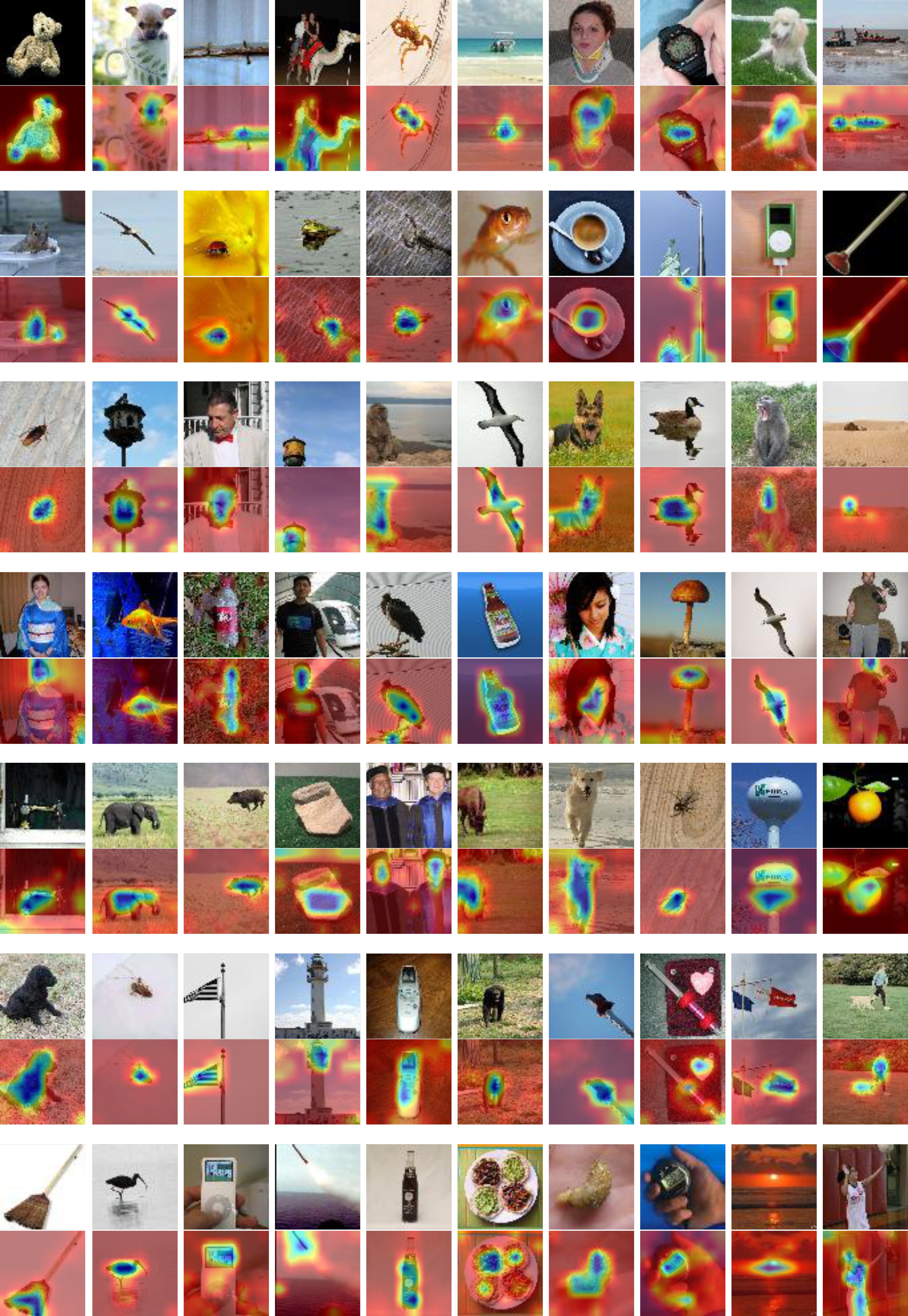}
    \end{center}
\caption{Sample Attention Maps of ViT trained using PatchRot on validation set of Tiny-ImageNet. Odd row: Input and Even row: Attention Map}
\label{Fig:attmap_all}
\end{figure}

We show more attention maps on the validation set of Tiny-ImageNet dataset in Figure \ref{Fig:attmap_all}.

\end{document}